\documentclass{article}

\usepackage[preprint]{neurips_2026}

\usepackage[utf8]{inputenc} 
\usepackage[T1]{fontenc}    
\usepackage{hyperref}       
\usepackage{url}            
\usepackage{booktabs}       
\usepackage{amsfonts}       
\usepackage{multirow}
\usepackage{nicefrac}       
\usepackage{microtype}      
\usepackage{xcolor}         
\usepackage{amsmath}
\usepackage{graphicx}       
\usepackage{subcaption}     
\usepackage{enumitem}

\title{What to Ignore, What to React: Visually Robust RL Fine-Tuning of VLA Models}

\author{
Yuanfang Peng$^{1,2}$ \quad
Jingjing Fu$^{2}$ \quad
Chuheng Zhang$^{2}$ \quad
Li Zhao$^{2}$ \quad
Jiang Bian$^{2}$  \\  
\textbf{Mingyu Liu$^{3}$ \quad
Ling Zhang$^{2}$\thanks{Corresponding authors.} \quad
Jun Zhang$^{1}$\footnotemark[1]  \quad Rui Wang$^{2}$\footnotemark[1]} \\
$^{1}$Hong Kong University of Science and Technology \\
$^{2}$Microsoft Research Asia  \\
$^{3}$Zhejiang University  \\
\texttt{ypengbx@connect.ust.hk},
\texttt{eejzhang@ust.hk} \\
\texttt{\{ds.dashu, jiang.bian.prc, wrui0920\}@gmail.com}\\
\texttt{\{chuhengzhang, lizo, zhangling\}@microsoft.com}
}

\begin{document}

\maketitle

\begin{abstract}

Reinforcement learning (RL) fine-tuning has shown promise for Vision-Language-Action (VLA) models in robotic manipulation, but deployment-time visual shifts pose practical challenges.
A key difficulty is that standard task rewards supervise task success, but offer limited guidance on whether a visual change is task-irrelevant or changes the behavior required for manipulation.
We propose PAIR-VLA (Paired Action Invariance \& Sensitivity for Visually Robust VLA), an RL fine-tuning framework to address this difficulty by adding two auxiliary objectives over paired visual variants during PPO optimization:
an invariance term that reduces the discrepancy between action distributions for a task-preserving pair (e.g., different distractors), and a sensitivity objective that encourages separable action distributions for a task-altering pair (e.g., target object in a different pose).
Together, these objectives turn visual variants from mere observation diversity into behavior-level guidance on policy responses during RL fine-tuning.
Because the auxiliary objectives are used only during RL fine-tuning, deployment will use the same policy architecture, thus incurring no additional inference-time cost.
We evaluate on ManiSkill3 across two representative VLA architectures, OpenVLA and $\pi_{0.5}$, under diverse out-of-distribution visual shifts including unseen distractors, texture changes, target object pose variation, viewpoint shifts, and lighting changes.
Our method consistently improves over standard PPO, achieving average absolute improvements of 16.62\% on $\pi_{0.5}$ and 9.10\% on OpenVLA.  
Notably, ablations further show generalization across visual shifts: invariance guidance learned from distractor and texture variants transfers to target-pose and lighting shifts, while adding sensitivity guidance on target-pose variants further improves robustness to nuisance shifts, highlighting the broader transferability of behavior-level RL guidance.

\end{abstract}

\section{Introduction}

Vision-Language-Action (VLA) models, pretrained on internet-scale vision-language data and diverse robot demonstrations~\citep{o2024openx, khazatsky2024droid}, have emerged as a promising paradigm for general-purpose robotic manipulation~\citep{brohan2023rt2visionlanguageactionmodelstransfer,octomodelteam2024octoopensourcegeneralistrobot, black2026pi0visionlanguageactionflowmodel}. 
However, pretraining does not guarantee  robustness under visual shifts at deployment.
Policies may remain sensitive to nuisance changes such as lighting, scene texture, or the presence of distractor objects~\citep{burns2023makespretrainedvisualrepresentations, xing2025shortcutlearninggeneralistrobot}, even when these changes do not alter the underlying manipulation objective.
These variations are challenging not only because they change the visual input, but also because they have different consequences for action: some should leave the intended manipulation unchanged, while others require the policy to adjust its behavior.
Such action-dependent visual shifts remain a challenge when adapting VLA policies to downstream manipulation tasks.

Reinforcement learning (RL) fine-tuning has recently been explored as a post-training approach for improving VLA policies on downstream manipulation tasks~\citep{li2025simplevlarlscalingvlatraining, yu2025rlinf, liu2026rlbringvlageneralization, chen2026pitextttrlonlinerlfinetuning, zang2026rlinfvlaunifiedefficientframework}.
Unlike supervised fine-tuning (SFT) on fixed demonstrations, RL optimizes expected task rewards through interaction, allowing the policy to adapt beyond the demonstration distribution.
However, visual robustness remains an important challenge during RL fine-tuning, as policies can still exhibit limited generalization under deployment-time visual shifts~\citep{liu2026rlbringvlageneralization, zhang2025robustvla}.
This raises the question of how visual variation should be incorporated into the RL fine-tuning process.


Existing approaches often improve generalization by increasing the diversity of training observations. 
In visual RL, image-level augmentations such as random cropping or color perturbation have been widely used to improve perceptual robustness~\citep{kostrikov2021imageaugmentationneedregularizing,laskin2020reinforcementlearningaugmenteddata, yarats2021masteringvisualcontinuouscontrol}. In robotics and sim-to-real learning, domain randomization is commonly applied over textures, lighting, camera poses, object placements, and distractor configurations~\cite{tobin2017domain, tremblay2018training, mehta2020active}.
These techniques are effective in many settings, but they primarily expose the policy to more varied observations. 
However, in  RL-based VLA fine-tuning, observation diversity alone does not inform the policy how its actions should respond to different types of scene changes, i.e., which changes should be ignored, and which changes should alter the required manipulation.  
This behavioral distinction is central to visual generalization in RL fine-tuning. 
The same policy should be invariant to changes that do not alter the required manipulation, while being sensitive to changes that do. 
We therefore study how to guide VLA policies during RL so that scene-level variation is aligned with action-level responses. 

Our method, PAIR-VLA (Paired Action Invariance \& Sensitivity for Visually Robust VLA), encourages two complementary behaviors: maintaining action consistency under task-irrelevant variations, and adapting actions when scene changes alter the required manipulation. Specifically, during RL training, we construct two perturbed views of each observation: a task-preserving view, where nuisance visual factors are altered while preserving the underlying task state, and a task-altering view, where the target object is perturbed. 
We then add two auxiliary objectives to PPO~\citep{schulman2017proximal}: an invariance objective that aligns the action distributions induced by the original and task-preserving views, and a sensitivity objective that separates those induced by the original and task-altering views.
This turns visual variants from mere observation diversity into explicit behavior-level supervision, specifying which scene changes the policy should ignore and which should induce a different action response.

We evaluate our approach on ManiSkill3~\citep{tao2024maniskill3} with two representative VLA architectures: the autoregressive OpenVLA~\citep{kim2024openvla} and the flow-based $\pi_{0.5}$~\citep{intelligence2025pi_}. 
Across a broad suite of out-of-distribution visual shifts, including unseen distractors, texture changes, target object pose variation, viewpoint shifts, and altered lighting conditions, our method consistently outperforms standard PPO fine-tuning, improving average success rates by 16.62\% on $\pi_{0.5}$ and 9.1\% on OpenVLA. 
Moreover, the policy generalizes to unseen lighting changes even though lighting was not used to construct the auxiliary objectives. 
Ablations show that invariance guidance transfers to other unseen shifts, such as target pose, while adding sensitivity guidance further improves robustness to nuisance changes, highlighting the broader generalization enabled by behavior-level RL guidance.

In summary, our contributions are as follows:
\begin{itemize}[leftmargin=*]
    \item We formulate visual generalization in VLA RL fine-tuning as behavior-level guidance over policy responses to visual shifts, which can reduce the reliance on extensive observation diversity during training and improve transfer to out-of-distribution scenes.
    \item We propose PAIR-VLA, a PPO-compatible invariance--sensitivity framework that implements behavior-level guidance over action distributions using paired visual variants. The invariance objective stabilizes actions under task-preserving visual changes, while the sensitivity objective encourages adaptive actions when task-relevant scene state changes. These auxiliary objectives are applied only during RL fine-tuning, leaving the deployed policy and inference cost unchanged.
\item We validate our method on ManiSkill3 with OpenVLA and $\pi_{0.5}$, showing consistent OOD improvements over PPO across diverse visual shifts. Ablations demonstrate that invariance and sensitivity guidance enable behavior-level RL generalization to unseen shifts.
\end{itemize}

\section{Related Work}

\paragraph{Online RL Fine-Tuning for VLA Models.}
Recent work has increasingly explored online RL as a post-training stage for VLA models, enabling policies pretrained or fine-tuned by supervised learning to further improve through environment interaction and task-completion rewards.
FLaRe~\citep{hu2025flare} introduces stabilization techniques for PPO fine-tuning of multi-task BC-pretrained policies; SimpleVLA~\citep{li2025simplevlarlscalingvlatraining} applies GRPO to OpenVLA-OFT~\citep{kim2025finetuningvisionlanguageactionmodelsoptimizing}; and RLinf~\citep{yu2025rlinf,zang2026rlinfvlaunifiedefficientframework} provides a unified, efficient framework for scalable RL training of VLA models. Meanwhile, RL4VLA~\citep{liu2026rlbringvlageneralization} evaluates OpenVLA generalization under PPO, GRPO, and DPO, finding PPO more effective for VLAs than LLM-derived methods such as DPO and GRPO, which motivates our use of PPO as the base algorithm. More recently, $\pi_{\mathrm{RL}}$~\citep{chen2026pitextttrlonlinerlfinetuning} studies online RL for flow-based VLA models such as $\pi_0$ and $\pi_{0.5}$. 
However, RL4VLA also reports that RL performs comparably to SFT on vision tasks, hypothesizing that neither training paradigm induces visual robustness beyond the visual randomness present during training. 
Rather than relying on data diversity alone during training, we augment PPO with action-distribution objectives that directly shape policy responses to different visual changes, improving robustness beyond the training distribution.








\paragraph{Visual Robustness in Robot Manipulation.}

Multiple lines of research have studied how to improve visual robustness
in robot learning and reduce the influence of nuisance visual variation.
One direct strategy is to modify observations at inference time. For
example, prior work learns masks to suppress visual distractors
~\citep{grooten2023madi}, or uses external segmentation and image editing
modules, such as inpainting, to remove irrelevant visual content before
feeding observations to the policy~\citep{
song2026overcoming, wu2025pcd, hancock2024runtimeobservationinterventionsmake}.
While such methods can reduce the effect of distractors at deployment,
they require extra modules to identify or remove irrelevant visual content at inference time, and do not directly train the policy itself to decide
which visual changes should affect its behavior.

A more common training-time strategy is to expose the policy to more
diverse visual observations. In visual RL, image-level augmentations such
as random cropping, color perturbation, and random shifts  
are widely used when learning
policies from pixels
~\citep{
kostrikov2021imageaugmentationneedregularizing,
laskin2020reinforcementlearningaugmenteddata,
yarats2021masteringvisualcontinuouscontrol}.
In sim-to-real robot learning, domain randomization varies simulation
parameters such as textures, lighting, camera poses, object placements,
and distractor configurations to reduce the visual gap between simulation
and deployment
~\citep{tobin2017domain, tremblay2018training, mehta2020active}.
Generative image editing, including inpainting and image synthesis, has also been used to create more realistic or diverse training observations~\citep{yu2023scaling, ho2021retinagan}.
These methods improve robustness through broader observation diversity, but do not specify how the policy's actions should respond to different types of scene changes.

Another line of work regularizes visual representations across
perturbed views. These methods encourage invariance across perturbed observations through
representation learning objectives, including contrastive learning,
bisimulation-based state abstraction, and consistency regularization in
feature or value space~\citep{
zhang2020learning, chen20contrastive, laskin2020curl,
bertoin2022look, sun2025salienceinvariantconsistentpolicylearning, yang2025invariance}.
Related approaches use information bottleneck to 
learn only task-relevant visual representations~\citep{james2022q,pacelli2020learningtaskdrivencontrolpolicies}.
These approaches provide useful
representation-level regularization, but invariance in latent space does
not necessarily guarantee the desired behavior at the action-distribution
level.

Recent work has begun to incorporate robustness-aware objectives into RL
post-training. Closest to our setting, RobustVLA~\citep{zhang2025robustvla} improves the resilience
of VLA policies to observation noise and action perturbations during
online RL training using Jacobian and smoothness regularization. Our setting differs in that we target broader scene-level visual shifts in robot manipulation, such as changes in distractor objects and target location.
Such shifts are not merely local corruptions, but can change whether the policy should preserve or adapt its manipulation behavior.
BiPS~\citep{zhang2025see} introduces
consistency and separation objectives during RL training of VLMs for
visual question answering. While we draw inspiration from this principle, VLA policy learning differs from VLM reasoning in the output space, the role of visual changes, and the policy architectures used to generate actions. We therefore ground the consistency--separation signal in manipulation consequences, using paired visual variants to keep policy behavior stable under task-preserving changes and adapt under task-altering changes. This allows visual variations to serve as behavior-level guidance during RL fine-tuning, while leaving the deployed VLA policy architecture unchanged.


\section{Preliminaries}

\textbf{Language-Conditioned Manipulation.} We model language-conditioned robotic manipulation as a partially observable Markov decision process (POMDP), defined as a tuple
$\mathcal{M}=(\mathcal{S}, \mathcal{A}, \mathcal{P}, R, \mathcal{O}, \gamma)$, augmented with a language instruction space $\mathcal{L}$. 
At the beginning of each episode, the robot receives a natural language instruction $l\in\mathcal{L}$,  which remains fixed throughout the episode. At each time step $t$,  the robot observes $o_t \in \mathcal{O}$
as a partial view of the underlying state $s_t \in \mathcal{S}$, 
and outputs an action $a_t\sim\pi_\theta(\cdot \mid o_t, l)$, where $\pi_\theta$ denotes the policy.
The environment transitions follow $s_{t+1}\sim \mathcal{P}(\cdot\mid s_t,a_t)$.

\textbf{Vision-Language-Action Models.} We instantiate the policy $\pi_\theta(\cdot \mid o, l)$ using VLA models, which map visual observations and language instructions to robot actions. 
We consider two representative architectues of VLA models: 
autoregressive models (e.g., OpenVLA~\citep{kim2024openvla}), and flow-matching models (e.g., $\pi_{0.5}$~\citep{intelligence2025pi_}). Our method applies to both architectures.


\textbf{RL fine-tuning and PPO.}
We use Proximal Policy Optimization (PPO)~\citep{schulman2017proximal} as our RL fine-tuning baseline, due to its empirical stability and effectiveness in continuous control settings~\cite{liu2026rlbringvlageneralization, chen2026pitextttrlonlinerlfinetuning}. Given an old policy $\pi_{\theta_{\mathrm{old}}}$, PPO optimizes the clipped surrogate objective
\begin{equation}
\mathcal{J}_{\mathrm{PPO}}(\theta) =
\mathbb{E}_{\tau \sim \pi_{\theta_{\mathrm{old}}}}
\left[
\frac{1}{T}\sum_{t=1}^{T}
\min\left(
w_t(\theta)\hat{A}_t,\,
\mathrm{clip}\left(w_t(\theta),1-\epsilon,1+\epsilon\right)\hat{A}_t
\right)
\right],
\end{equation}
where
$w_t(\theta)=\frac{\pi_\theta(a_t \mid o_t, l)}{\pi_{\theta_{\mathrm{old}}}(a_t \mid o_t, l)}$ is the likelihood ratio between the updated and old policies, $\epsilon$ is the clipping parameter, and $\hat{A}_t$ denotes the advantage estimate computed using Generalized Advantage Estimation (GAE)~\citep{schulman2015high}.

PPO can be applied directly to autoregressive VLAs via their explicit action log-likelihoods. 
Flow-matching VLAs, however, generate actions through iterative ODE denoising and lack tractable action log probabilities, making the standard PPO likelihood-ratio objective difficult to apply directly.
We follow $\pi_{\mathrm{RL}}$~\cite{chen2026pitextttrlonlinerlfinetuning}, which converts the ODE denoising process into an SDE and formulates a two-layer MDP. This yields tractable log-likelihoods and enables PPO updates.

\section{Visual Robustness via Invariance and Sensitivity Objectives}\label{sec:sec_method}


\subsection{Problem Formulation}\label{sec:problem}

Our goal is to learn a policy that is robust to task-irrelevant visual variations while remaining sensitive to task-relevant ones. 
Consider a policy $a\sim \pi_\theta(\cdot \mid o, l)$ conditioned on an observation $o$ and a language instruction $l$. 
We define a visually robust policy as one satisfying two properties: 
(1) \emph{task-preserving invariance}: the action distribution is approximately preserved under task-irrelevant visual perturbations, including the presence of distractor objects, changes in viewpoint, and changes in background appearance such as table textures;
(2) \emph{task-altering sensitivity}: the action distribution changes meaningfully under perturbations to the task-relevant objects.

To operationalize these properties during RL fine-tuning, we introduce controlled perturbations of the input observation by constructing two types of perturbed views. 
The first is a \emph{task-preserving view}, 
which removes distractor objects and varies background appearance factors such as textures and colors 
while preserving the underlying task state and required manipulation. 
The second is a \emph{task-altering view}, 
which modifies task-relevant properties of the target object (e.g., object pose) while preserving its semantic identity.

These views instantiate the task-irrelevant and task-relevant visual changes used by our auxiliary objectives.

\subsection{Invariance and Sensitivity Objectives}

Using the task-preserving and task-altering views, we add two complementary auxiliary objectives to PPO training. Both are formulated as 
KL divergences between the action distributions induced by the original and perturbed observations. 
The invariance objective aligns the policy distribution under the original and task-preserving views, encouraging consistency under visual changes that preserve the required manipulation. 
The sensitivity objective separates the policy distribution under the original and task-altering views, encouraging the policy to respond when the required manipulation changes.



\textbf{Invariance objective.}
To enforce task-preserving invariance, we encourage the policy to produce consistent action distributions under visual changes that do not alter the required manipulation.
Let $\tilde{o}_t^{\mathrm{prev}}$ denote the task-preserving view of $o_t$.
We then minimize the KL divergence between the action distributions induced by the original and task-preserving views:
\begin{equation}
    \mathcal{L}_{\mathrm{inv}}(\theta)
    = \mathbb{E}_{(o_t, l)}\Big[ D_{\mathrm{KL}}\big(
    \pi_\theta(\cdot \mid o_t, l)
    \;\|\;
    \text{sg}[\pi_\theta(\cdot \mid \tilde{o}_t^{\mathrm{prev}}, l)]
    \big)\Big],
\end{equation}
where $\mathrm{sg}[\cdot]$ denotes the stop-gradient operator. It treats the task-preserving-view distribution as a fixed target to stabilize training.


\textbf{Sensitivity objective.}

The invariance objective specifies which visual changes should leave the policy distribution unchanged, but provides no signal for which changes should alter it. 
In manipulation, task-relevant changes such as target object pose can require different grasps, motions, or placements. 
We therefore introduce a sensitivity objective that provides the complementary signal, explicitly encouraging the policy distribution to change under task-altering perturbations.


Let $\tilde{o}_t^{\mathrm{alt}}$ denote the task-altering view of $o_t$.
We then maximize the KL divergence between the action distributions induced by the original and task-altering views:
\begin{equation}
    \mathcal{L}_{\mathrm{sens}}(\theta)
    = - \mathbb{E}_{(o_t, l)}\Big[ \min \left(
    c,\;
    D_{\mathrm{KL}}\big(
    \pi_\theta(\cdot \mid o_t, l)
    \;\|\;
    \text{sg}[\pi_\theta(\cdot \mid \tilde{o}_t^{\mathrm{alt}}, l)]
    \big)
    \right)\Big],
\end{equation}
where $c$ is a clipping threshold that prevents unbounded divergence growth from destablizing PPO training. 

\begin{figure}[t]
  \centering
  \includegraphics[width=0.72\linewidth]{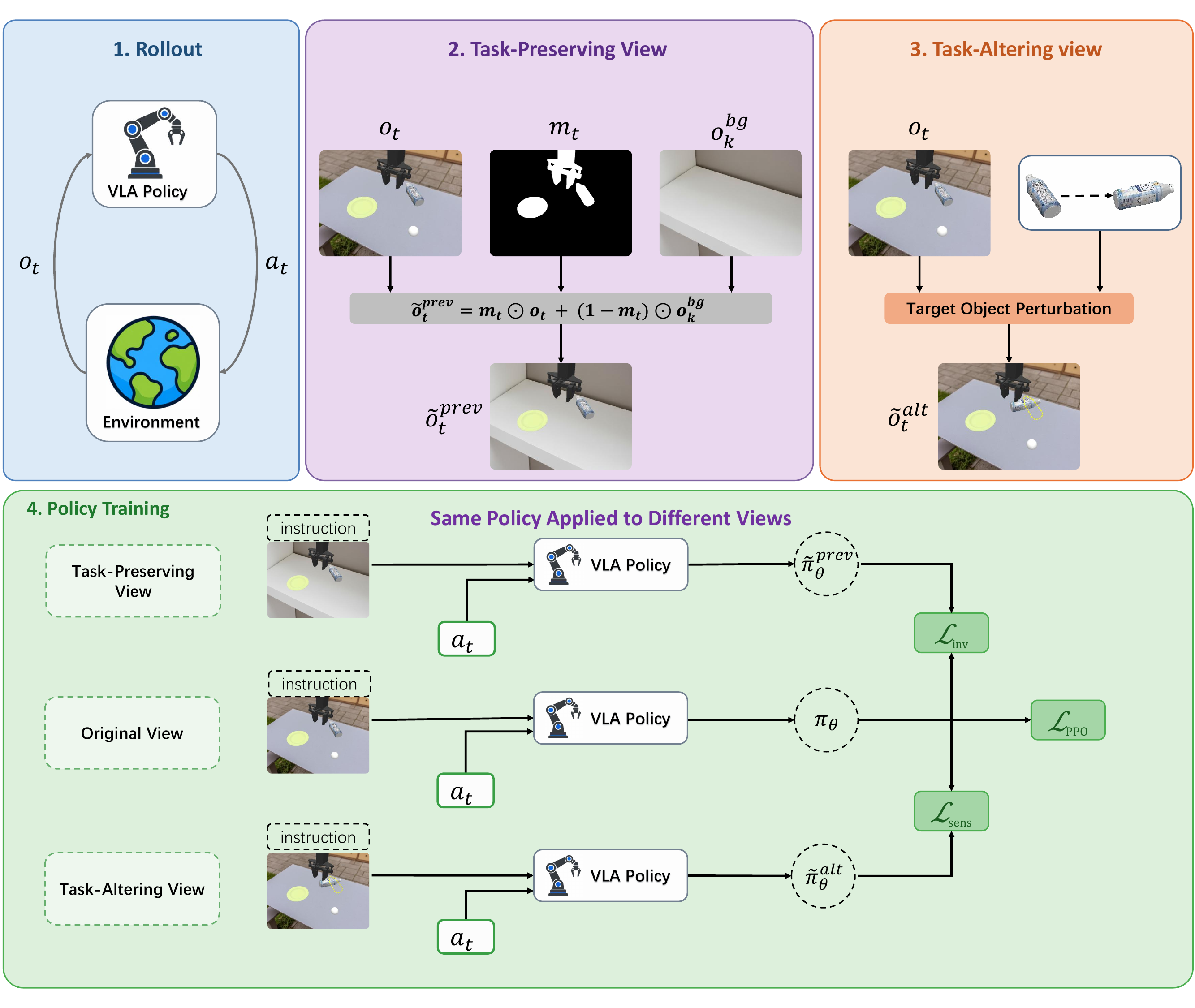}
  \caption{Overview of the visually robust RL fine-tuning framework. At each environment step, the VLA policy receives an observation $o_t$ from the environment and samples an action $a_t$. Based on $o_t$, the task-preserving view is constructed by removing distractors and changing background appearance, while the task-altering view is constructed by perturbing the pose of the target object.
  During policy training, the original view and paired variants are separately passed through the same VLA policy $\pi_\theta$ to compute their action distributions.
  The invariance and sensitivity objectives are computed over these action distributions, 
  and optimized as auxiliary objectives during PPO training.
  }
  \label{fig:framework}
\end{figure}

\textbf{Augmented PPO objective.} 
Combining the invariance and sensitivity objectives with the standard PPO objective, we obtain the overall training objective:
\begin{equation}
    \mathcal{L}(\theta) = \mathcal{L}_{\mathrm{PPO}}(\theta)
    + \alpha \mathcal{L}_{\mathrm{inv}}(\theta)
    + \beta \mathcal{L}_{\mathrm{sens}}(\theta),
\end{equation}
where $\alpha, \beta >0$ are coefficients controlling the relative strength of each auxiliary objective. 
The two terms are complementary: $\mathcal{L}_{\mathrm{inv}}$ discourages the policy from responding to task-irrelevant visual changes, while $\mathcal{L}_{\mathrm{sens}}$ encourages responsiveness to changes in task-relevant object state. 
Together, they provide explicit action-level guidance on which visual changes should preserve the policy's distribution and which should alter it. Figure~\ref{fig:framework} summarizes the overall fine-tuning framework.



\subsection{Augmented View Construction}

We now describe how we implement the two perturbed views used by the invariance and sensitivity objectives.

\textbf{Task-preserving view.} The task-preserving view alters task-irrelevant visual factors while preserving the underlying task state and required manipulation. 
In our main construction, we instantiate this view by removing distractors and changing background appearance through foreground-background compositing. 
Task-relevant foreground content from the current observation is composited with background snapshots from different scene configurations. Segmentation masks for foreground extraction are obtained directly from the simulator via object identifiers; in real-world settings, they can be approximated by an off-the-shelf segmentation model such as SAM 3~\citep{carion2025sam}.

Prior to training, we pre-render a set of $K$ background snapshots $\{o_k^{\mathrm{bg}}\}_{k=1}^K$ by rendering each scene configuration with all objects set to invisible. During training, for each observation $o_t$, 
we define a binary segmentation mask $m_t$ that takes value $1$ for pixels belonging to the robot, target object, or receptacle, and $0$ otherwise.
We then sample a background snapshot $o_k$ uniformly from the pre-rendered set, and construct the task-preserving view as
\begin{equation}
\tilde{o}_t^{\mathrm{prev}}= m_t \odot o_t +(1-m_t)\odot o_k^{\mathrm{bg}}.
\end{equation}
This operation removes distractors and replaces background appearance while preserving task-relevant content.

To evaluate whether the proposed invariance objective can improve generalization to unseen viewpoints, we also construct an alternative task-preserving view for viewpoint variation in Section~\ref{sec:exp-camera}.

\textbf{Task-altering view.}  
The task-altering view modifies task-relevant visual content while preserving the semantic identity of the target object. 
We construct this view by perturbing the pose of the target object. 
Specifically, at each training step, we sample a translation perturbation from a Gaussian distribution and a rotation perturbation from a categorical distribution, apply them to the target object, and re-render the scene to obtain $\tilde{o}_t^{\mathrm{alt}}$. This produces target object pose variations that can change the required grasp, motion, or placement. 
In simulation, we implement this by
directly setting the object pose through the simulator API. 

\section{Experiments}\label{sec:sec_results}

We evaluate whether our method improves the visual robustness of RL-fine-tuned VLA models across visual shifts.
The experiments are designed to answer four questions:
(1) whether the proposed auxiliary objectives improve OOD generalization over standard PPO,
(2) whether they improve RL fine-tuning efficiency,
(3) how each objective
contributes to performance, and
(4) whether the  proposed method generalizes to other task-preserving visual changes such as viewpoint shifts.

\subsection{Experimental Setup}\label{subsec:exp_setup}

\textbf{Models and Training.}
For the autoregressive VLA, we use OpenVLA~\citep{kim2024openvla}; for the flow-matching VLA, we use $\pi_{0.5}$~\citep{intelligence2025pi_}. 
For each backbone, we first obtain an SFT checkpoint and use it to initialize both the PPO baseline and our method. We then fine-tune both models under the same PPO settings, where our method augments PPO with the proposed auxiliary objectives.
Details
of the SFT procedure, PPO settings, and auxiliary-objective hyperparameters are
provided in Appendix~\ref{app:sft-train}.

\textbf{Benchmark.} We conduct experiments in the Maniskill3~\citep{tao2024maniskill3} simulator and focus on a representative \emph{pick-and-place} task with visual distractors, where the agent must place a target object onto a receptacle (a plate in our setup), despite the presence of distractors.

To systematically evaluate visual robustness, we design a set of out-of-distribution (OOD) scenarios that vary visual factors beyond those seen in training.
During training, target objects, distractors, and table textures are sampled from predefined training sets. To prevent overfitting to specific spatial configurations, at the beginning of each episode, the initial positions of the target object, distractors and receptacle are randomly sampled from a $6\times6$ grid of discrete positions on the table surface. 

At test time, we then assess generalization under OOD scenarios where the following visual factors differ from those seen during training: (1) \textit{unseen table texture}, where table textures are sampled from a held-out set; (2) \textit{unseen lighting}, where scene lighting is varied using unseen configurations; (3) \textit{unseen poses}, where the target object pose is sampled from translation and rotation distributions that differ from those used during training; and (4) \textit{unseen clutter}, where the number of distractors is increased, with half of the distractors sampled from a held-out object set. The corresponding train/test splits are provided in {Appendix~\ref{app:benchmark}}.

\textbf{Metrics.} We evaluate performance using task success rate, defined as the fraction of evaluation episodes in which the agent successfully completes the task. For each run, success rate is computed over 128 evaluation episodes, and we report the mean across three independent runs for each task.


\subsection{Main Results}


\textbf{OOD generalization.}
Table~\ref{tab:maniskill-main-results} 
reports success rates on the four OOD visual generalization settings defined above: unseen table texture, unseen lighting, unseen target pose, and unseen clutter. The Distractor column averages over test settings with 2, 4, 6, and 8 distractors, which we analyze in more detail in Section~\ref{sec:analysis}. Across both OpenVLA and $\pi_{0.5}$, our method consistently improves over PPO in every OOD setting. On OpenVLA, our method improves the average OOD success rate from 77.90\% to 87.00\%, yielding a gain of 9.10 points. On $\pi_{0.5}$, the improvement is larger, increasing the average success rate from 46.25\% to 62.87\%, for a gain of 16.62 points. These results show that adding our auxiliary objectives to PPO improves visual generalization across both autoregressive and flow-based VLA backbones.


\begin{table}[t]
  \caption{OOD generalization of our method compared to PPO across four OOD scenarios on OpenVLA and $\pi_{0.5}$. Values are success rates (\%). The Clutter column reports the average over test settings with 2, 4, 6, and 8 distractors. $\Delta$ Avg. is computed relative to the PPO baseline.}
  \label{tab:maniskill-main-results}
  \centering
  {\small
  \setlength{\tabcolsep}{5.0pt}
  \renewcommand{\arraystretch}{1.14}
  \begin{tabular}{@{}llcccccc@{}}
    \toprule
    Model & Method &
      \multicolumn{6}{c}{OOD settings} \\
    \cmidrule(l){3-8}
    & &
      \shortstack{Table  Texture} &
      \shortstack{Lighting} &
      \shortstack{Target  Pose} &
      \shortstack{Clutter} &
      Avg. & $\Delta$ Avg. \\
    \midrule
    \multirow{2}{*}{OpenVLA} & PPO &
      86.98 & 72.14 & 83.59 & 68.88 & 77.90 & -- \\
    & \textbf{Ours} &
      \textbf{94.53} & \textbf{80.47} & \textbf{90.63} &
      \textbf{82.36} & \textbf{87.00} &
      \textcolor{green!50!black}{\textbf{+9.10}} \\
    \midrule
    \multirow{2}{*}{$\pi_{0.5}$} & PPO &
      63.54 & 28.54 & 56.46 & 36.46 & 46.25 & -- \\
    & \textbf{Ours} &
      \textbf{80.21} & \textbf{51.67} & \textbf{69.38} &
      \textbf{50.21} & \textbf{62.87} &
      \textcolor{green!50!black}{\textbf{+16.62}} \\
    \bottomrule
  \end{tabular}}
\end{table}

\begin{figure}[t]
  \centering
  \begin{subfigure}[t]{0.45\linewidth}
    \centering
    \includegraphics[width=\linewidth]{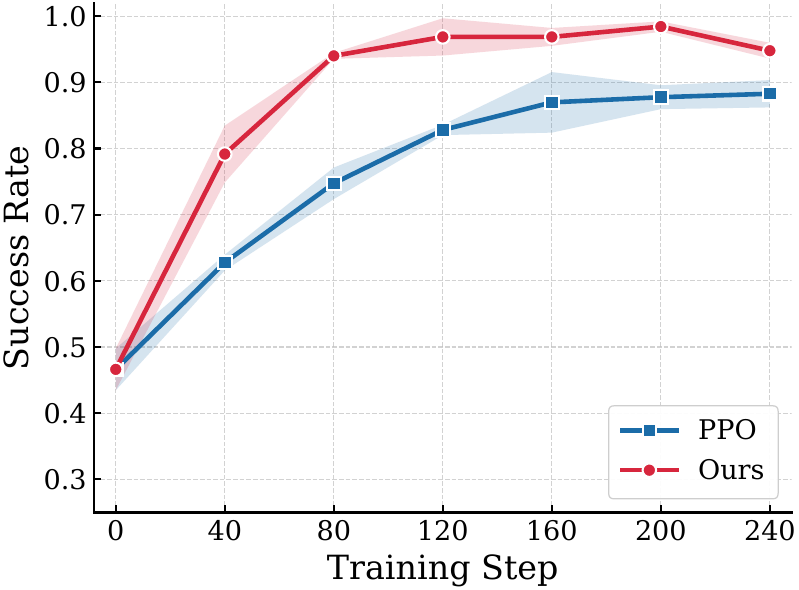}
    \caption{ID scenario with 1 distractor}
    \label{fig:se-train}
  \end{subfigure}
  \hfill
  \begin{subfigure}[t]{0.45\linewidth}
    \centering
    \includegraphics[width=\linewidth]{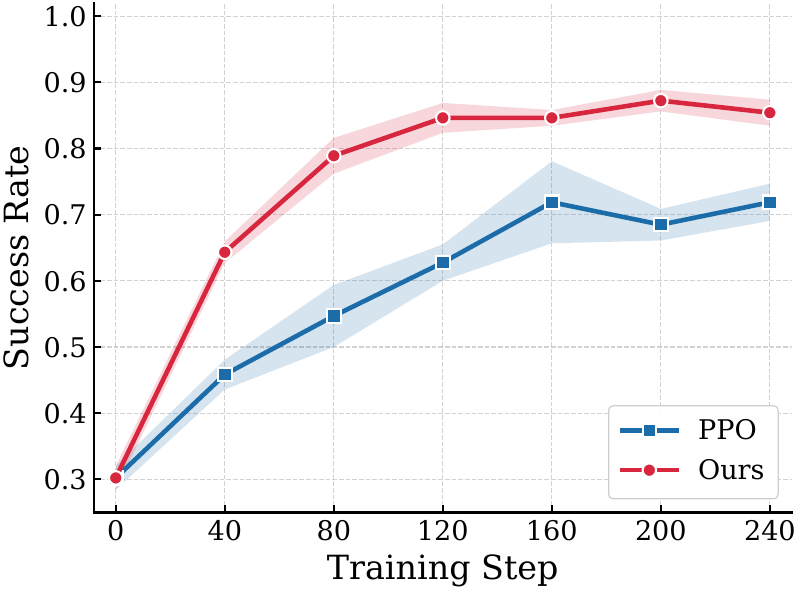}
    \caption{OOD scenario with 4 distractors.}
    \label{fig:se-ood}
  \end{subfigure}
  \caption{\textbf{RL fine-tuning efficiency on Maniskill3 with OpenVLA.} Success
  rate versus training step for our method and PPO on (a) an ID scenario and (b) an OOD clutter scenario. Solid lines show
  the mean over three seeds; shaded regions show the standard deviation.
  Our method converges substantially faster and reaches a higher plateau in both
  settings.}
  \label{fig:sample-efficiency}
\end{figure}

\textbf{RL Fine-Tuning Efficiency.}
To further assess whether our method improves the fine-tuning efficiency of PPO, we periodically evaluate OpenVLA checkpoints obtained at different training steps on both an in-distribution (ID) scenario (Fig.~\ref{fig:se-train}) and an OOD scenario (Fig.~\ref{fig:se-ood}). In the ID scenario, the scene contains one distractor sampled from the same object set used during training. In the OOD scenario, the scene contains four distractors, half sampled from the training object set and half from a held-out object set.

In the ID scenario, our method reaches a success rate of 90\% within 80 training steps, whereas PPO requires roughly 240 steps to reach the same level, yielding an approximate $3\times$ improvement in fine-tuning efficiency. The advantage also holds in the OOD scenario. These results indicate that our auxiliary objectives improve not only final OOD robustness but also the efficiency of PPO fine-tuning.



\subsection{Ablation Study}

We ablate the two auxiliary objectives in our method, the invariance objective
$\mathcal{L}_{\mathrm{inv}}$ and the sensitivity objective
$\mathcal{L}_{\mathrm{sens}}$,  using the
$\pi_{0.5}$ backbone on Maniskill3. Table~\ref{tab:ablation} reports OOD
success rates under the same four OOD settings as in 
Table~\ref{tab:maniskill-main-results}.
Adding $\mathcal{L}_{\mathrm{inv}}$ alone yields a large average gain of
$+12.61$ points over PPO, with particularly strong improvements under unseen
lighting ($+16.04$) and unseen clutter ($+13.54$). $\mathcal{L}_{\mathrm{sens}}$
alone gives a smaller but consistent gain of $+1.73$ points, but combining it
with $\mathcal{L}_{\mathrm{inv}}$ achieves the best results across all four OOD
settings, improving over PPO by $+16.62$ points on average. This suggests that
the invariance objective provides the main robustness signal, while the
sensitivity objective further improves robustness when used jointly.



\begin{table}[t]
\caption{Ablation of the two auxiliary objectives using the $\pi_{0.5}$ backbone. Values are success rates (\%). The Clutter column reports the average over test settings with 2, 4, 6, and 8 distractors. $\Delta$ Avg. is computed relative to the PPO baseline.}
  \label{tab:ablation}
  \centering
  {\small
  \setlength{\tabcolsep}{5.0pt}
  \renewcommand{\arraystretch}{1.14}
  \begin{tabular}{@{}lcccccc@{}}
    \toprule
    Method &
      \multicolumn{6}{c}{OOD settings} \\
    \cmidrule(l){2-7}
    &
      \shortstack{Table Texture} &
      \shortstack{Lighting} &
      \shortstack{Target Pose} &
      \shortstack{Clutter} &
      Avg. & $\Delta$ Avg. \\
    \midrule
    PPO &
      63.54 & 28.54 & 56.46 & 36.46 & 46.25 & -- \\
    PPO w/ $\mathcal{L}_{\mathrm{inv}}$ &
      72.92 & 44.58 & 67.92 & 50.00 & 58.86 &
      \textcolor{green!50!black}{+12.61} \\
    PPO w/ $\mathcal{L}_{\mathrm{sens}}$ &
      63.96 & 33.54 & 53.33 & 41.10 & 47.98 &
      \textcolor{green!50!black}{+1.73} \\
    \textbf{Ours} &
      \textbf{80.21} & \textbf{51.67} & \textbf{69.38} &
      \textbf{50.21} & \textbf{62.87} &
      \textcolor{green!50!black}{\textbf{+16.62}} \\
    \bottomrule
  \end{tabular}}
\end{table}


\subsection{Analysis}\label{sec:analysis}

\paragraph{OOD generalization under increasing clutter levels.}

\begin{figure}[t]
  \centering
  \begin{subfigure}[t]{0.45\linewidth}
    \centering
    \includegraphics[width=\linewidth]{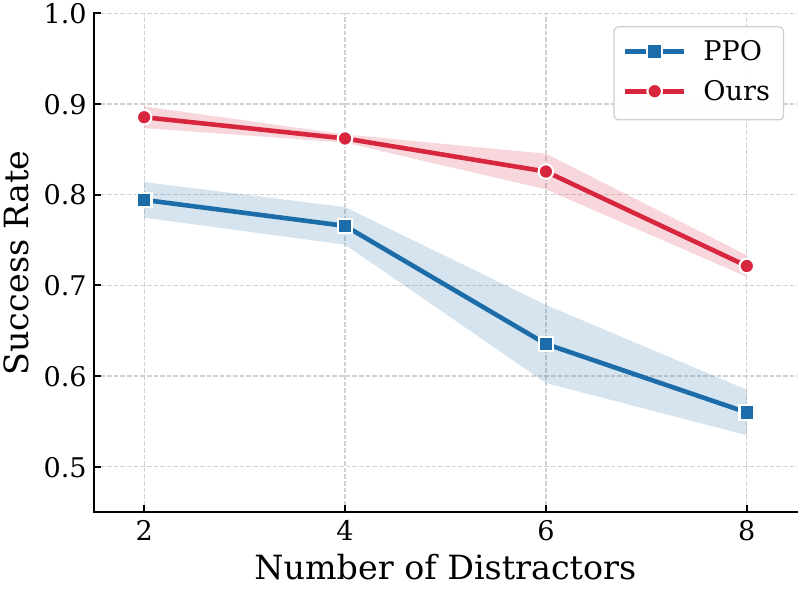}
    \caption{OpenVLA.}
    \label{fig:ext-openvla}
  \end{subfigure}
  \hfill
  \begin{subfigure}[t]{0.45\linewidth}
    \centering
    \includegraphics[width=\linewidth]{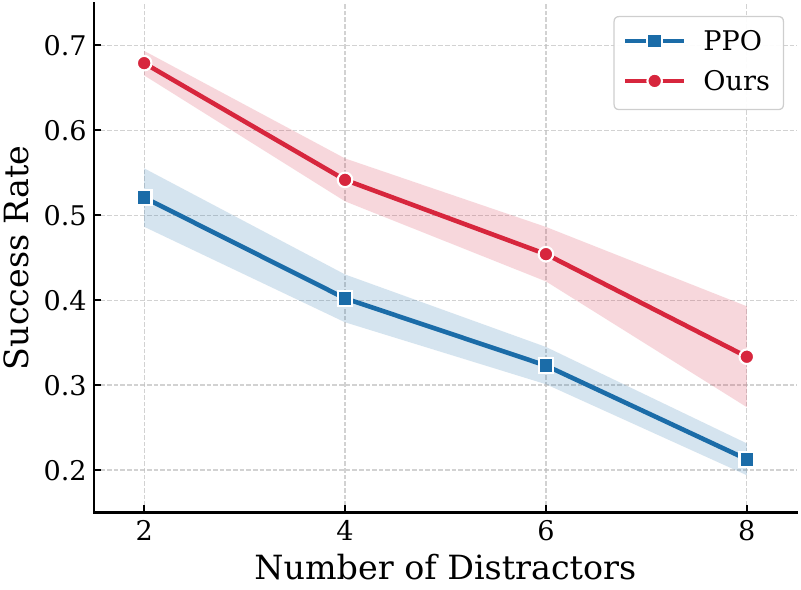}
    \caption{$\pi_{0.5}$.}
    \label{fig:ext-pi05}
  \end{subfigure}
    \caption{\textbf{OOD generalization under increasing clutter levels.}
    Success rate with 2--8 distractors for (a) OpenVLA and (b) $\pi_{0.5}$; half of the distractors in each setting are sampled from a held-out object set.}
    \label{fig:ood-extrapolation}
\end{figure}

We further analyze OOD generalization under varying clutter levels by evaluating OpenVLA and $\pi_{0.5}$ with 2 to 8 distractors; in each setting, half of the distractors are sampled from a held-out object set (Figures~\ref{fig:ext-openvla} and~\ref{fig:ext-pi05}).
As expected, success rates decrease for both PPO and our method as the number of unseen distractors increases. However, our method consistently maintains higher success rates across all tested clutter levels, showing a more robust degradation profile. In the most challenging setting with 8 distractors, our method achieves 72\% on OpenVLA and 33\% on $\pi_{0.5}$, compared to 56\% and 21\% for PPO, respectively.


\begin{figure}[t]
  \centering
  \begin{minipage}[t]{0.48\linewidth}
    \centering
    \includegraphics[width=\linewidth]{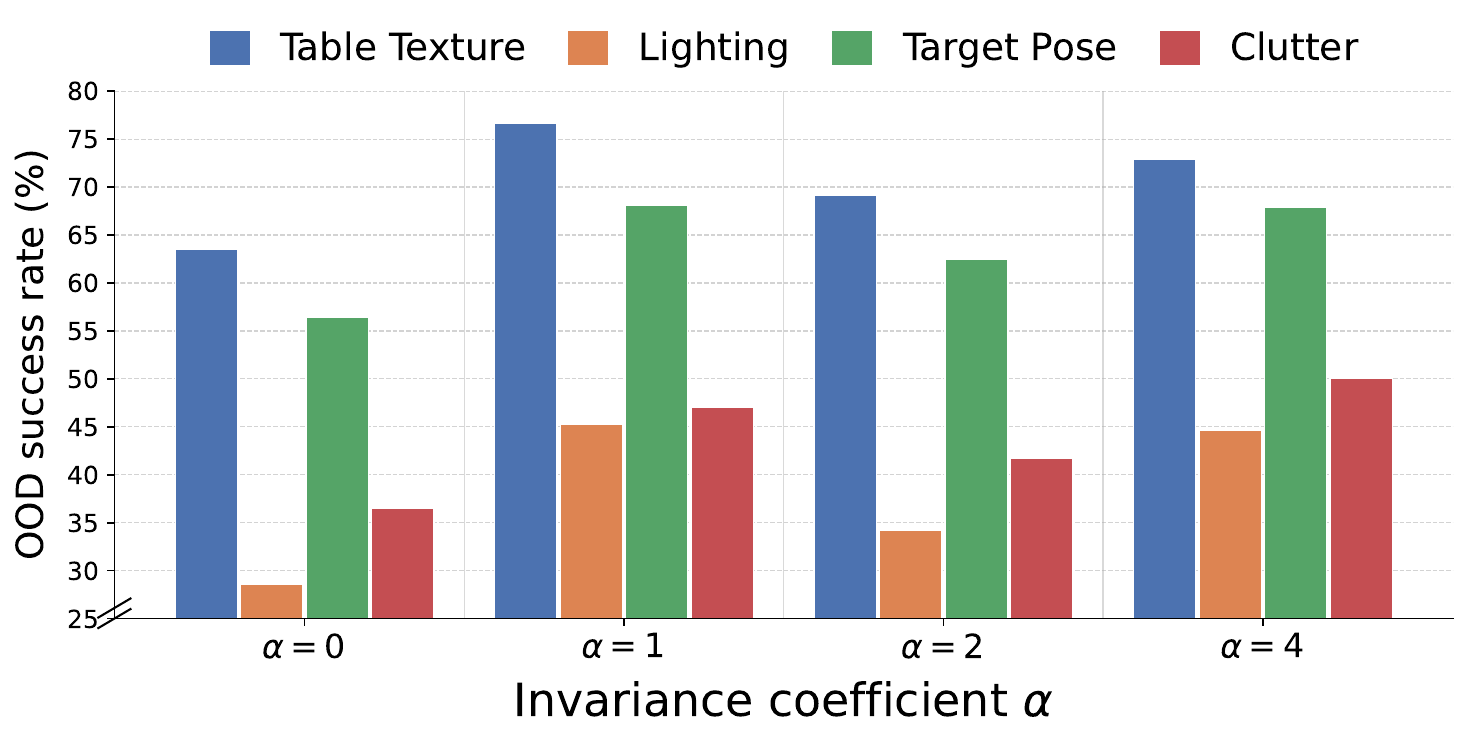}
    \captionof{figure}{\textbf{Effect of the invariance coefficient $\alpha$.}
    OOD success rates (\%) on ManiSkill3 with the $\pi_{0.5}$ backbone for
    $\alpha \in \{0,1,2,4\}$, with the sensitivity objective disabled
    ($\beta=0$). Here, $\alpha=0$ recovers the PPO baseline. Nonzero values of $\alpha$ consistently improve OOD performance, with
$\alpha=1$ achieving the best overall results and gains remaining stable across
the tested invariance weights.}
    \label{fig:ablate-alpha}
  \end{minipage}
  \hfill
  \begin{minipage}[t]{0.48\linewidth}
    \centering
    \includegraphics[width=\linewidth]{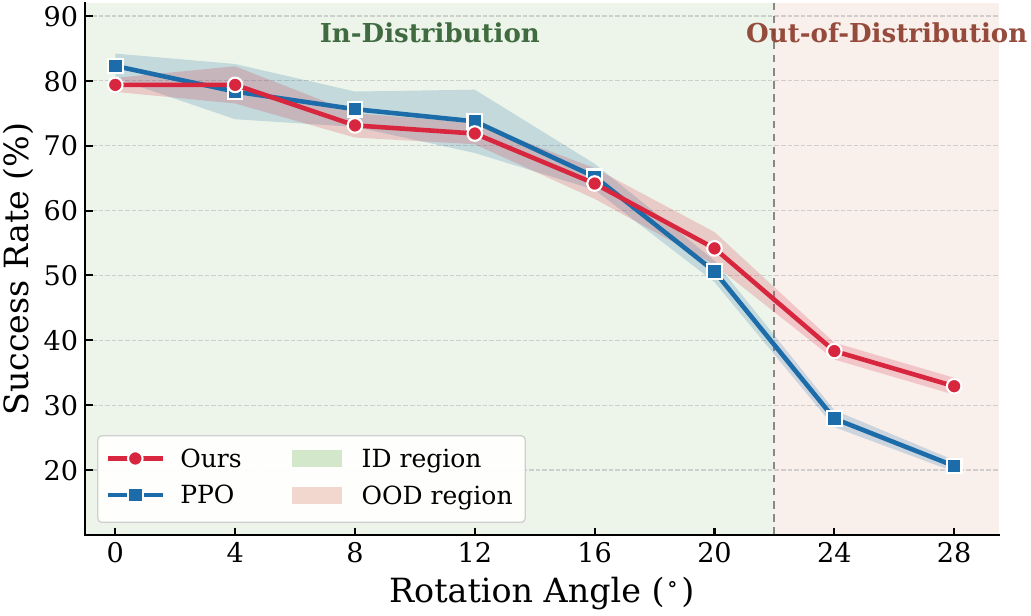}
    \captionof{figure}{\textbf{OOD extrapolation to unseen camera poses with
    $\pi_{0.5}$.} Success rate versus camera rotation angle.
    Green and salmon shading mark the training range $[0^\circ,20^\circ]$ and unseen angles $\{24^\circ,28^\circ\}$, respectively.
    Lines denote the mean over three seeds, with bands showing one standard deviation.
    Our method matches PPO within
    the training viewpoint range while significantly outperforming it under
    unseen camera poses.}
    \label{fig:rotation-ood-extrapolation}
  \end{minipage}
\end{figure}

\textbf{Effect of the Invariance Coefficient $\alpha$.}
Since the invariance objective provides the dominant robustness gain in the
ablation study, we further examine how its weight affects OOD generalization.
We sweep $\alpha \in \{0,1,2,4\}$ using the $\pi_{0.5}$ backbone while disabling
the sensitivity objective by setting $\beta=0$, so that $\alpha=0$ recovers the
PPO baseline. Figure~\ref{fig:ablate-alpha} reports success rates under the four
OOD settings: unseen table texture, unseen lighting, unseen target pose, and
unseen clutter. Across all settings, enabling the invariance objective improves
over PPO. The best performance is obtained at $\alpha=1$, which yields gains of
more than 10 points in each OOD setting and more than 15 points under unseen
lighting, although lighting is not directly varied in the task-preserving view construction used for the invariance objective. Performance remains strong at $\alpha=4$, indicating that the robustness gains are stable across different invariance weights.


\subsection{Generalization to Unseen Camera Viewpoints}\label{sec:exp-camera}

We further evaluate generalization to unseen camera viewpoints using an alternative task-preserving view for the invariance objective, where the same scene state is re-rendered from a different camera pose while preserving object poses, robot state, and the task goal. The task-altering view used for the sensitivity objective remains unchanged. During RL fine-tuning, the policy sees camera rotations within
$[0^\circ,20^\circ]$ at $4^\circ$ increments. At test time, we evaluate
rotations from $0^\circ$ to $28^\circ$ at the same $4^\circ$ increments, with
$\{24^\circ,28^\circ\}$ forming the OOD range.



Figure~\ref{fig:rotation-ood-extrapolation} reports success rates across camera viewpoints. In the ID range, our method achieves $70.36\%$ success, comparable to $70.97\%$ for PPO. In the OOD range, our method substantially outperforms PPO, achieving $38.34\%$ versus $27.92\%$ at $24^\circ$, and $32.92\%$ versus $20.63\%$ at $28^\circ$. This shows that, when instantiated with viewpoint-based task-preserving views, our method improves extrapolation to OOD camera poses while preserving in-distribution performance.

\section{Discussion \& Conclusion}\label{sec:sec_discussion}

We proposed PAIR-VLA, a visually robust RL fine-tuning framework for VLA models. 
The framework augments PPO with two auxiliary objectives that encourage invariance to task-irrelevant visual changes and sensitivity to task-relevant ones.
On ManiSkill3, our method consistently improves OOD success rates over PPO on both OpenVLA and $\pi_{0.5}$ across diverse visual shifts, including unseen table textures, target-pose shifts, and increasing visual clutter.
It improves generalization to unseen lighting conditions, even though lighting changes are not used to construct the paired views for our auxiliary objectives.
An additional camera-pose experiment further shows improved generalization to unseen camera viewpoints.
The method also improves RL fine-tuning efficiency, achieving higher in-distribution and OOD success rates with fewer PPO updates. 
A limitation of our current study is that evaluation is conducted in simulation, and task-preserving view construction relies on ground-truth object masks from the simulator.
Future work could study whether robustness learned in simulation transfers to real-world deployment, and examine how segmentation masks estimated by off-the-shelf models affect paired-view construction and the resulting robustness gains.
Importantly, both paired-view construction and the auxiliary objectives are used only during RL fine-tuning, so the deployed policy requires no additional inference-time modules.



\newpage
\bibliographystyle{unsrt}
\bibliography{reference}

@misc{zang2026rlinfvlaunifiedefficientframework,
      title={RLinf-VLA: A Unified and Efficient Framework for Reinforcement Learning of Vision-Language-Action Models}, 
      author={Hongzhi Zang and Mingjie Wei and Si Xu and Yongji Wu and Zhen Guo and Yuanqing Wang and Hao Lin and Peihong Wang and Liangzhi Shi and Yuqing Xie and Zhexuan Xu and Zhihao Liu and Kang Chen and Wenhao Tang and Quanlu Zhang and Weinan Zhang and Chao Yu and Yu Wang},
      year={2026},
      eprint={2510.06710},
      archivePrefix={arXiv},
      primaryClass={cs.RO},
      url={https://arxiv.org/abs/2510.06710}, 
}

@misc{chen2026pitextttrlonlinerlfinetuning,
      title={$\pi_\texttt{RL}$: Online RL Fine-tuning for Flow-based Vision-Language-Action Models}, 
      author={Kang Chen and Zhihao Liu and Tonghe Zhang and Zhen Guo and Si Xu and Hao Lin and Hongzhi Zang and Xiang Li and Quanlu Zhang and Zhaofei Yu and Guoliang Fan and Tiejun Huang and Yu Wang and Chao Yu},
      year={2026},
      eprint={2510.25889},
      archivePrefix={arXiv},
      primaryClass={cs.LG},
      url={https://arxiv.org/abs/2510.25889}, 
}

@misc{li2025simplevlarlscalingvlatraining,
      title={SimpleVLA-RL: Scaling VLA Training via Reinforcement Learning}, 
      author={Haozhan Li and Yuxin Zuo and Jiale Yu and Yuhao Zhang and Zhaohui Yang and Kaiyan Zhang and Xuekai Zhu and Yuchen Zhang and Tianxing Chen and Ganqu Cui and Dehui Wang and Dingxiang Luo and Yuchen Fan and Youbang Sun and Jia Zeng and Jiangmiao Pang and Shanghang Zhang and Yu Wang and Yao Mu and Bowen Zhou and Ning Ding},
      year={2025},
      eprint={2509.09674},
      archivePrefix={arXiv},
      primaryClass={cs.RO},
      url={https://arxiv.org/abs/2509.09674}, 
}

@misc{liu2026rlbringvlageneralization,
      title={What Can RL Bring to VLA Generalization? An Empirical Study}, 
      author={Jijia Liu and Feng Gao and Bingwen Wei and Xinlei Chen and Qingmin Liao and Yi Wu and Chao Yu and Yu Wang},
      year={2026},
      eprint={2505.19789},
      archivePrefix={arXiv},
      primaryClass={cs.LG},
      url={https://arxiv.org/abs/2505.19789}, 
}

@inproceedings{tobin2017domain,
  title={Domain randomization for transferring deep neural networks from simulation to the real world},
  author={Tobin, Josh and Fong, Rachel and Ray, Alex and Schneider, Jonas and Zaremba, Wojciech and Abbeel, Pieter},
  booktitle={2017 IEEE/RSJ international conference on intelligent robots and systems (IROS)},
  pages={23--30},
  year={2017},
  organization={IEEE}
}

@article{yang2025invariance,
  title={Invariance Co-training for Robot Visual Generalization},
  author={Yang, Jonathan and Finn, Chelsea and Sadigh, Dorsa},
  journal={arXiv preprint arXiv:2512.05230},
  year={2025}
}

@article{yu2023scaling,
  title={Scaling robot learning with semantically imagined experience},
  author={Yu, Tianhe and Xiao, Ted and Stone, Austin and Tompson, Jonathan and Brohan, Anthony and Wang, Su and Singh, Jaspiar and Tan, Clayton and Peralta, Jodilyn and Ichter, Brian and others},
  journal={arXiv preprint arXiv:2302.11550},
  year={2023}
  }

@misc{kostrikov2021imageaugmentationneedregularizing,
      title={Image Augmentation Is All You Need: Regularizing Deep Reinforcement Learning from Pixels}, 
      author={Ilya Kostrikov and Denis Yarats and Rob Fergus},
      year={2021},
      eprint={2004.13649},
      archivePrefix={arXiv},
      primaryClass={cs.LG},
      url={https://arxiv.org/abs/2004.13649}, 
}

@inproceedings{hu2025flare,
  title={Flare: Achieving masterful and adaptive robot policies with large-scale reinforcement learning fine-tuning},
  author={Hu, Jiaheng and Hendrix, Rose and Farhadi, Ali and Kembhavi, Aniruddha and Mart{\'\i}n-Mart{\'\i}n, Roberto and Stone, Peter and Zeng, Kuo-Hao and Ehsani, Kiana},
  booktitle={2025 IEEE International Conference on Robotics and Automation (ICRA)},
  pages={3617--3624},
  year={2025},
  organization={IEEE}
}

@article{schulman2017proximal,
  title={Proximal policy optimization algorithms},
  author={Schulman, John and Wolski, Filip and Dhariwal, Prafulla and Radford, Alec and Klimov, Oleg},
  journal={arXiv preprint arXiv:1707.06347},
  year={2017}
}

@article{zhang2025see,
  title={See Less, See Right: Bi-directional Perceptual Shaping For Multimodal Reasoning},
  author={Zhang, Shuoshuo and Zhang, Yizhen and Fu, Jingjing and Song, Lei and Bian, Jiang and Yang, Yujiu and Wang, Rui},
  journal={arXiv preprint arXiv:2512.22120},
  year={2025}
}

@article{tao2024maniskill3,
  title={Maniskill3: Gpu parallelized robotics simulation and rendering for generalizable embodied ai},
  author={Tao, Stone and Xiang, Fanbo and Shukla, Arth and Qin, Yuzhe and Hinrichsen, Xander and Yuan, Xiaodi and Bao, Chen and Lin, Xinsong and Liu, Yulin and Chan, Tse-kai and others},
  journal={arXiv preprint arXiv:2410.00425},
  year={2024}
}

@article{kim2024openvla,
  title={Openvla: An open-source vision-language-action model},
  author={Kim, Moo Jin and Pertsch, Karl and Karamcheti, Siddharth and Xiao, Ted and Balakrishna, Ashwin and Nair, Suraj and Rafailov, Rafael and Foster, Ethan and Lam, Grace and Sanketi, Pannag and others},
  journal={arXiv preprint arXiv:2406.09246},
  year={2024}
}

@article{intelligence2025pi_,
  title={$\pi_{0.5}$: A Vision-Language-Action Model with Open-World Generalization},
  author={Intelligence, Physical and Black, Kevin and Brown, Noah and Darpinian, James and Dhabalia, Karan and Driess, Danny and Esmail, Adnan and Equi, Michael and Finn, Chelsea and Fusai, Niccolo and others},
  journal={arXiv preprint arXiv:2504.16054},
  year={2025}
}

@article{yu2025rlinf,
  title={Rlinf: Flexible and efficient large-scale reinforcement learning via macro-to-micro flow transformation},
  author={Yu, Chao and Wang, Yuanqing and Guo, Zhen and Lin, Hao and Xu, Si and Zang, Hongzhi and Zhang, Quanlu and Wu, Yongji and Zhu, Chunyang and Hu, Junhao and others},
  journal={arXiv preprint arXiv:2509.15965},
  year={2025}
}

@InProceedings{chen20contrastive,
  title = 	 {A Simple Framework for Contrastive Learning of Visual Representations},
  author =       {Chen, Ting and Kornblith, Simon and Norouzi, Mohammad and Hinton, Geoffrey},
  booktitle = 	 {Proceedings of the 37th International Conference on Machine Learning},
  pages = 	 {1597--1607},
  year = 	 {2020},
  editor = 	 {III, Hal Daumé and Singh, Aarti},
  volume = 	 {119},
  series = 	 {Proceedings of Machine Learning Research},
  month = 	 {13--18 Jul},
  publisher =    {PMLR},
  url = 	 {https://proceedings.mlr.press/v119/chen20j.html}
}

@inproceedings{laskin2020curl,
  title={Curl: Contrastive unsupervised representations for reinforcement learning},
  author={Laskin, Michael and Srinivas, Aravind and Abbeel, Pieter},
  booktitle={International conference on machine learning},
  pages={5639--5650},
  year={2020},
  organization={PMLR}
}

@article{zhang2020learning,
  title={Learning invariant representations for reinforcement learning without reconstruction},
  author={Zhang, Amy and McAllister, Rowan and Calandra, Roberto and Gal, Yarin and Levine, Sergey},
  journal={arXiv preprint arXiv:2006.10742},
  year={2020}
}

@article{grooten2023madi,
  title={Madi: Learning to mask distractions for generalization in visual deep reinforcement learning},
  author={Grooten, Bram and Tomilin, Tristan and Vasan, Gautham and Taylor, Matthew E and Mahmood, A Rupam and Fang, Meng and Pechenizkiy, Mykola and Mocanu, Decebal Constantin},
  journal={arXiv preprint arXiv:2312.15339},
  year={2023}
}

@article{bertoin2022look,
  title={Look where you look! saliency-guided q-networks for generalization in visual reinforcement learning},
  author={Bertoin, David and Zouitine, Adil and Zouitine, Mehdi and Rachelson, Emmanuel},
  journal={Advances in neural information processing systems},
  volume={35},
  pages={30693--30706},
  year={2022}
}

@article{song2026overcoming,
  title={Overcoming Visual Clutter in Vision Language Action Models via Concept-Gated Visual Distillation},
  author={Song, Sangmim and Kodagoda, Sarath and Carmichael, Marc and Thiyagarajan, Karthick},
  journal={arXiv preprint arXiv:2603.10340},
  year={2026}
}

@article{schulman2015high,
  title={High-dimensional continuous control using generalized advantage estimation},
  author={Schulman, John and Moritz, Philipp and Levine, Sergey and Jordan, Michael and Abbeel, Pieter},
  journal={arXiv preprint arXiv:1506.02438},
  year={2015}
}

@article{carion2025sam,
  title={Sam 3: Segment anything with concepts},
  author={Carion, Nicolas and Gustafson, Laura and Hu, Yuan-Ting and Debnath, Shoubhik and Hu, Ronghang and Suris, Didac and Ryali, Chaitanya and Alwala, Kalyan Vasudev and Khedr, Haitham and Huang, Andrew and others},
  journal={arXiv preprint arXiv:2511.16719},
  year={2025}
}

@misc{brohan2023rt2visionlanguageactionmodelstransfer,
      title={RT-2: Vision-Language-Action Models Transfer Web Knowledge to Robotic Control}, 
      author={Anthony Brohan and Noah Brown and Justice Carbajal and Yevgen Chebotar and Xi Chen and Krzysztof Choromanski and Tianli Ding and Danny Driess and Avinava Dubey and Chelsea Finn and Pete Florence and Chuyuan Fu and Montse Gonzalez Arenas and Keerthana Gopalakrishnan and Kehang Han and Karol Hausman and Alexander Herzog and Jasmine Hsu and Brian Ichter and Alex Irpan and Nikhil Joshi and Ryan Julian and Dmitry Kalashnikov and Yuheng Kuang and Isabel Leal and Lisa Lee and Tsang-Wei Edward Lee and Sergey Levine and Yao Lu and Henryk Michalewski and Igor Mordatch and Karl Pertsch and Kanishka Rao and Krista Reymann and Michael Ryoo and Grecia Salazar and Pannag Sanketi and Pierre Sermanet and Jaspiar Singh and Anikait Singh and Radu Soricut and Huong Tran and Vincent Vanhoucke and Quan Vuong and Ayzaan Wahid and Stefan Welker and Paul Wohlhart and Jialin Wu and Fei Xia and Ted Xiao and Peng Xu and Sichun Xu and Tianhe Yu and Brianna Zitkovich},
      year={2023},
      eprint={2307.15818},
      archivePrefix={arXiv},
      primaryClass={cs.RO},
      url={https://arxiv.org/abs/2307.15818}, 
}

@misc{octomodelteam2024octoopensourcegeneralistrobot,
      title={Octo: An Open-Source Generalist Robot Policy}, 
      author={Octo Model Team and Dibya Ghosh and Homer Walke and Karl Pertsch and Kevin Black and Oier Mees and Sudeep Dasari and Joey Hejna and Tobias Kreiman and Charles Xu and Jianlan Luo and You Liang Tan and Lawrence Yunliang Chen and Pannag Sanketi and Quan Vuong and Ted Xiao and Dorsa Sadigh and Chelsea Finn and Sergey Levine},
      year={2024},
      eprint={2405.12213},
      archivePrefix={arXiv},
      primaryClass={cs.RO},
      url={https://arxiv.org/abs/2405.12213}, 
}

@article{black2026pi0visionlanguageactionflowmodel,
  title={$\pi_0 $: A Vision-Language-Action Flow Model for General Robot Control},
  author={Black, Kevin and Brown, Noah and Driess, Danny and Esmail, Adnan and Equi, Michael and Finn, Chelsea and Fusai, Niccolo and Groom, Lachy and Hausman, Karol and Ichter, Brian and others},
  journal={arXiv preprint arXiv:2410.24164},
  year={2024}
}

@inproceedings{o2024openx,
  title={Open x-embodiment: Robotic learning datasets and rt-x models: Open x-embodiment collaboration 0},
  author={O’Neill, Abby and Rehman, Abdul and Maddukuri, Abhiram and Gupta, Abhishek and Padalkar, Abhishek and Lee, Abraham and Pooley, Acorn and Gupta, Agrim and Mandlekar, Ajay and Jain, Ajinkya and others},
  booktitle={2024 IEEE International Conference on Robotics and Automation (ICRA)},
  pages={6892--6903},
  year={2024},
  organization={IEEE}
}

@article{khazatsky2024droid,
  title={Droid: A large-scale in-the-wild robot manipulation dataset},
  author={Khazatsky, Alexander and Pertsch, Karl and Nair, Suraj and Balakrishna, Ashwin and Dasari, Sudeep and Karamcheti, Siddharth and Nasiriany, Soroush and Srirama, Mohan Kumar and Chen, Lawrence Yunliang and Ellis, Kirsty and others},
  journal={arXiv preprint arXiv:2403.12945},
  year={2024}
}

@misc{burns2023makespretrainedvisualrepresentations,
      title={What Makes Pre-Trained Visual Representations Successful for Robust Manipulation?}, 
      author={Kaylee Burns and Zach Witzel and Jubayer Ibn Hamid and Tianhe Yu and Chelsea Finn and Karol Hausman},
      year={2023},
      eprint={2312.12444},
      archivePrefix={arXiv},
      primaryClass={cs.CV},
      url={https://arxiv.org/abs/2312.12444}, 
}

@misc{kim2025finetuningvisionlanguageactionmodelsoptimizing,
      title={Fine-Tuning Vision-Language-Action Models: Optimizing Speed and Success}, 
      author={Moo Jin Kim and Chelsea Finn and Percy Liang},
      year={2025},
      eprint={2502.19645},
      archivePrefix={arXiv},
      primaryClass={cs.RO},
      url={https://arxiv.org/abs/2502.19645}, 
}

@article{zhang2025robustvla,
  title={RobustVLA: Robustness-aware reinforcement post-training for vision-language-action models},
  author={Zhang, Hongyin and Zhang, Shuo and Jin, Junxi and Zeng, Qixin and Li, Runze and Wang, Donglin},
  journal={arXiv preprint arXiv:2511.01331},
  year={2025}
}

@misc{xing2025shortcutlearninggeneralistrobot,
      title={Shortcut Learning in Generalist Robot Policies: The Role of Dataset Diversity and Fragmentation}, 
      author={Youguang Xing and Xu Luo and Junlin Xie and Lianli Gao and Hengtao Shen and Jingkuan Song},
      year={2025},
      eprint={2508.06426},
      archivePrefix={arXiv},
      primaryClass={cs.RO},
      url={https://arxiv.org/abs/2508.06426}, 
}

@article{wu2025pcd,
  title={Policy contrastive decoding for robotic foundation models},
  author={Wu, Shihan and Luo, Xu and Zhang, Ji and Xie, Junlin and Song, Jingkuan and Shen, Heng Tao and Gao, Lianli},
  journal={arXiv preprint arXiv:2505.13255},
  year={2025}
}

@misc{laskin2020reinforcementlearningaugmenteddata,
      title={Reinforcement Learning with Augmented Data}, 
      author={Michael Laskin and Kimin Lee and Adam Stooke and Lerrel Pinto and Pieter Abbeel and Aravind Srinivas},
      year={2020},
      eprint={2004.14990},
      archivePrefix={arXiv},
      primaryClass={cs.LG},
      url={https://arxiv.org/abs/2004.14990}, 
}

@inproceedings{mehta2020active,
  title={Active domain randomization},
  author={Mehta, Bhairav and Diaz, Manfred and Golemo, Florian and Pal, Christopher J and Paull, Liam},
  booktitle={Conference on Robot Learning},
  pages={1162--1176},
  year={2020},
  organization={PMLR}
}

@inproceedings{tremblay2018training,
  title={Training deep networks with synthetic data: Bridging the reality gap by domain randomization},
  author={Tremblay, Jonathan and Prakash, Aayush and Acuna, David and Brophy, Mark and Jampani, Varun and Anil, Cem and To, Thang and Cameracci, Eric and Boochoon, Shaad and Birchfield, Stan},
  booktitle={Proceedings of the IEEE conference on computer vision and pattern recognition workshops},
  pages={969--977},
  year={2018}
}

@misc{yarats2021masteringvisualcontinuouscontrol,
      title={Mastering Visual Continuous Control: Improved Data-Augmented Reinforcement Learning}, 
      author={Denis Yarats and Rob Fergus and Alessandro Lazaric and Lerrel Pinto},
      year={2021},
      eprint={2107.09645},
      archivePrefix={arXiv},
      primaryClass={cs.AI},
      url={https://arxiv.org/abs/2107.09645}, 
}

@inproceedings{ho2021retinagan,
  title={Retinagan: An object-aware approach to sim-to-real transfer},
  author={Ho, Daniel and Rao, Kanishka and Xu, Zhuo and Jang, Eric and Khansari, Mohi and Bai, Yunfei},
  booktitle={2021 IEEE International Conference on Robotics and Automation (ICRA)},
  pages={10920--10926},
  year={2021},
  organization={IEEE}
}

@misc{sun2025salienceinvariantconsistentpolicylearning,
      title={Salience-Invariant Consistent Policy Learning for Generalization in Visual Reinforcement Learning}, 
      author={Jingbo Sun and Songjun Tu and Qichao Zhang and Ke Chen and Dongbin Zhao},
      year={2025},
      eprint={2502.08336},
      archivePrefix={arXiv},
      primaryClass={cs.AI},
      url={https://arxiv.org/abs/2502.08336}, 
}

@misc{hancock2024runtimeobservationinterventionsmake,
      title={Run-time Observation Interventions Make Vision-Language-Action Models More Visually Robust}, 
      author={Asher J. Hancock and Allen Z. Ren and Anirudha Majumdar},
      year={2024},
      eprint={2410.01971},
      archivePrefix={arXiv},
      primaryClass={cs.RO},
      url={https://arxiv.org/abs/2410.01971}, 
}

@misc{pacelli2020learningtaskdrivencontrolpolicies,
      title={Learning Task-Driven Control Policies via Information Bottlenecks}, 
      author={Vincent Pacelli and Anirudha Majumdar},
      year={2020},
      eprint={2002.01428},
      archivePrefix={arXiv},
      primaryClass={cs.LG},
      url={https://arxiv.org/abs/2002.01428}, 
}

@article{james2022q,
  title={Q-attention: Enabling efficient learning for vision-based robotic manipulation},
  author={James, Stephen and Davison, Andrew J},
  journal={IEEE Robotics and Automation Letters},
  volume={7},
  number={2},
  pages={1612--1619},
  year={2022},
  publisher={IEEE}
}

@article{hu2022lora,
  title={Lora: Low-rank adaptation of large language models.},
  author={Hu, Edward J and Shen, Yelong and Wallis, Phillip and Allen-Zhu, Zeyuan and Li, Yuanzhi and Wang, Shean and Wang, Liang and Chen, Weizhu and others},
  journal={Iclr},
  volume={1},
  number={2},
  pages={3},
  year={2022}
}

\newpage
\appendix

\section{SFT and RL Training Details}\label{app:sft-train}
\subsection{SFT Checkpoints}
The VLA models are pre-trained on large-scale demonstration data. However, they still struggle to perform the downstream tasks out-of-the-box. In that case, a SFT stage is required to warm-up the VLA models.We use the checkpoints of OpenVLA and $\pi_{0.5}$ models provided by RLinf\citep{yu2025rlinf}. These checkpoints are finetuned from their pre-trained models with around $16k$ demonstrations collected by motion planning. Even though these demonstrations are collected in environments where no distractor is placed on the table, we observed that models finetuned on such data still show non-trivial performance on our training task with one distractor placed on the table. We therefore use these checkpoints as initializations for RL training.

\subsection{RL Training Details}\label{app:rl-train}

\paragraph{LoRA finetuning}
For the OpenVLA model, we use Low-Rank Adaptation
(LoRA)\citep{hu2022lora} for RL finetuning to reduce the computation overhead. For the $\pi_{0.5}$ model, even though the recommended training recipe from $\pi_{RL}$\citep{chen2026pitextttrlonlinerlfinetuning} is to train the action expert only, we observed that the OOD performance of both PPO baseline and our method can be further improved. In that case, we also apply LoRA to the VLM backbone and full-finetune the action expert. For both OpenVLA and $\pi_{0.5}$, we choose LoRA rank $r=32$ and apply LoRA modules to all linear Layers.

\paragraph{Training and auxiliary-objective hyperparameters.}
Table~\ref{tab:hyperparams_single_task} reports the RL fine-tuning and
auxiliary-objective hyperparameters used for OpenVLA and $\pi_{0.5}$. PPO hyperparameters are shared between the PPO baseline and our
method for each backbone. Our method additionally uses the auxiliary-objective
coefficients $\alpha$ and $\beta$ for the invariance and sensitivity objectives,
respectively, as well as the sensitivity clipping parameter $c$. These
auxiliary hyperparameters are selected separately for OpenVLA and $\pi_{0.5}$
because the two policy parameterizations induce KL divergences on different
numerical scales. We note that $c$ is used only in the sensitivity objective and
is distinct from the PPO clipping parameter.

\paragraph{Compute Resources}
For both our method and the baseline, each experiment is conducted on 8$\times$H100 GPUs. For OpenVLA, the training process requires approximately 1 day to complete. For $\pi_{0.5}$, the training process requires approximately 3 days to complete.


\begin{table}[t]
  \caption{RL fine-tuning and auxiliary-objective hyperparameters for OpenVLA and
$\pi_{0.5}$. RL fine-tuning hyperparameters are shared by the PPO baseline and
our method for each backbone; auxiliary-objective hyperparameters are used only
by our method.}
  \label{tab:hyperparams_single_task}
  \centering
\begin{tabular}{lcc}
    \toprule
    \textbf{Parameter} & \textbf{OpenVLA} & $\boldsymbol{\pi_{0.5}}$ \\
    \midrule
    \multicolumn{3}{l}{\textit{RL fine-tuning}} \\
    RL train steps & 240 & 280 \\
    Global batch size & 640 & 5120 \\
    Update epochs & 1 & 5 \\
    Actor learning rate & $1{\times}10^{-4}$ & $7.91{\times}10^{-6}$ \\
    Critic learning rate & $3{\times}10^{-3}$ & $1.55{\times}10^{-4}$ \\
    Reward discount rate $\gamma$ & 0.99 & 0.99 \\
    GAE $\lambda$ & 0.95 & 0.95 \\
    PPO clip ratio $\epsilon$ & 0.2 & 0.2 \\
    \midrule
    \multicolumn{3}{l}{\textit{Rollout collection}} \\
    Interaction steps & 80 & 80 \\
    Parallel environments & 128 & 320 \\
    Rollout epochs & 1 & 1 \\
    \midrule
    \multicolumn{3}{l}{\textit{Auxiliary objectives}} \\
    Invariance coefficient $\alpha$ & 1 & 4 \\
    Sensitivity coefficient $\beta$ & 0.2 & 4 \\
    Sensitivity clip $c$ & 0.8 & 0.08 \\
    \midrule
    \multicolumn{3}{l}{\textit{Action generation}} \\
    Action prediction horizon $H$ & 1 & 8 \\
    Action replan horizon $H'$ & 1 & 5 \\
    Denoise steps & -- & 4 \\
    Noise level $\sigma$ (Flow-SDE) & -- & 0.5 \\
    \bottomrule
\end{tabular}
\end{table}

\section{Benchmark Details}~\label{app:benchmark}
We conduct experiments in the Maniskill3\citep{tao2024maniskill3} simulator and focus on a representative \emph{pick-and-place} task, where the agent is instructed to pick up the target object and place it on the plate. In this task, observations consist of a $480\times640$ third-person image, joint-poses and instructions. Actions are defined in delta end-effector poses. This task provide three kinds of rewards:
\begin{itemize}
  \item \emph{outcome reward:} give a reward of $+1$ is the task succeed and $0$ otherwise.
  \item \emph{is grasp reward:} give a reward of $+0.1$ if the robot grasp the target object and $0$ otherwise.
  \item \emph{consecutive grasp reward:} give a reward of $+0.1$ if the robot grasp the target object for few steps and $0$ otherwise.
\end{itemize}
\subsection{Table Textures and Objects}
We collect 21 different  table textures and 25 different categroies of objects, following the setup of RL4VLA\citep{liu2026rlbringvlageneralization}. The table textures are separated to 16 candidates for training environments and 5 candidates for evaluation environments. The objects are also splitted to the training set with 16 categories and the evaluation set with 9 categories. Only the table textures and objects belonging to the training set will be exposed to VLA models during SFT and RL training.
\subsection{Scene Lighting}
The scene lighting is composed of a ambient light, a key light and two fill lights. As we want to test how the visual robustness learned from other kinds of variations transfer to OOD lighting conditions, the parameters of the scene lighting are fixed during SFT and RL training. Instead, we define 20 OOD lighting conditions by vary these parameters, which is held-out from training.
\subsection{Object Position}
For training, the positions of the objects are sampled from a square area centered at [-0.16,0] and length of the half-edge is 7.5cm. This square area is discretized to $6\times6$ grids and the initial positions are sampled from these grid points. Orientation is sampled from a set $\{0, \frac{1}{4} \pi, \frac{1}{2}\pi, \pi \}$. For evaluation, the square area is extended outtoward to half-edge$\approx 10.5$cm. The initialization position of the objects are sampled from the outer border.
\subsection{Training and Evaluation Tasks}
To systematically investigate the visual OOD robustness, we build multiple tasks as follows:
\begin{itemize}
    \item \emph{Training:} At the beginning of each episode, two objects are sampled from the training objects and one table texture is sampled from the training table textures. One of the sampled objects is the target object while another is the distractor. The appearance of the plate is fixed. Positions of the objects and the plate is sampled from the $6\times6$ grid in the small square area. The lighting condition is also fixed in the training task.
    \item \emph{Unseen Table Texture:} The table texture is sampled from evaluation set. Other settings are the same as in \emph{Training}.
    \item \emph{Unseen Lighting:} The parameters of the lighting conditions are sampled the evaluation set with 20 different combinations. Other settings are the same as \emph{Training}.
    \item \emph{Unseen Pose} The position of the target object is sampled from outer border of the enlarged square area. Other settings are the same as \emph{Training}.
    \item \emph{Unseen Clutter} One object is sampled from the training set as the target object. $N$ objects are sampled as distractors, where half of the objects are from training set and another half from evaluation set. We use "Distractor-$N$" to denote the task with $N$ distractors. Other settings are the same as \emph{Training}.
\end{itemize}

\section{Broader Impacts}\label{app:impact}
Our work may have positive impact by improving the robustness and reliability of robot manipulation policies under visual shifts, which can reduce brittle failures in embodied AI systems such as assistive robots.

Potential negative impacts include unsafe deployment of visually robust policies in unvalidated real-world settings or misuse of improved robotic capabilities. We encourage proper and responsible use of our findings.





\end{document}